# Multi-Method Validation of Large Language Model Medical Translation Across High- and Low-Resource Languages

Chukwuebuka Anyaegbuna, MD; Eduardo Juan Perez Guerrero; Jerry Liu, MD; Timothy Keyes, PhD; April Liang, MD; Natasha Steele, MD, MPH; Stephen Ma, MD; Jonathan Chen, MD, PhD; Kevin Schulman, MD, MBA

**Corresponding Author:** Chukwuebuka Anyaegbuna, MD, 3180 Porter Drive, Palo Alto, CA 94304; gozirim@gmail.com; +1 628 309 3402

## Abstract

Language barriers affect 27.3 million U.S. residents with non-English language preference, yet professional medical translation remains costly and often unavailable. We evaluated four frontier large language models (GPT-5.1, Claude Opus 4.5, Gemini 3 Pro, Kimi K2) translating 22 medical documents into 8 languages spanning high-resource (Spanish, Chinese, Russian, Vietnamese), medium-resource (Korean, Arabic), and low-resource (Tagalog, Haitian Creole) categories using a five-layer validation framework. Across 704 translation pairs, all models achieved high semantic preservation (LaBSE greater than 0.92), with no significant difference between high- and low-resource languages (p = 0.066). Cross-model back-translation confirmed results were not driven by same-model circularity (delta = -0.0009). Inter-model concordance across four independently trained models was high (LaBSE: 0.946), and lexical borrowing analysis showed no correlation between English term retention and fidelity scores in low-resource languages (rho = +0.018, p = 0.82). These converging results suggest frontier LLMs preserve medical meaning across resource levels, with implications for language access in healthcare.

## Introduction

Access to medical information is fundamental to patient engagement and health outcomes. Studies encompassing over 74,000 patients have demonstrated that patient education significantly improves physiological, physical, and psychological outcomes while reducing medication use and healthcare utilization.1 Yet for the 27.3 million U.S. residents with non-English language preference, language barriers create substantial obstacles to accessing this information.2-4

The consequences are well-documented. Adults with limited English proficiency (LEP) report poorer health outcomes, including lower rates of glycemic control, higher rates of uncontrolled asthma, and increased odds of poorly controlled hypertension.3 They are significantly less likely to utilize outpatient visits and are three times as likely as the English-proficient population to be uninsured.3 Medical errors experienced by LEP individuals are more likely to cause physical harm compared to those experienced by English-proficient patients.4 As the Joint Commission notes in their Roadmap for Hospitals, "effective communication is now accepted as an essential component of quality care and patient safety."5

Federal law recognizes these disparities. Section 1557 of the Affordable Care Act and the National Standards for Culturally and Linguistically Appropriate Services (CLAS) mandate that healthcare organizations provide language access services.6,7 However, manual professional medical translation is costly and resource-intensive and is not always available in a timely or workflow-compatible manner during routine clinical care.8,9 As a result, professional translation is often applied selectively rather than comprehensively, with documented gaps for speakers of less common languages and for written materials generated during rapid clinical processes such as hospital discharge.10

Large language models have demonstrated remarkable capabilities in translation.11 Given their ready accessibility, these technologies could address critical gaps in medical information availability. However, medical translation presents unique challenges: terminology must be precise, instructions unambiguous, and errors can have serious consequences.12 Early evaluations suggest LLM translation quality varies by language: a recent study found ChatGPT and Google Translate performed comparably to professional translation for Spanish and Portuguese discharge instructions but had significantly more clinically significant errors for Haitian Creole.13 The reliability of LLMs across languages, particularly low-resource languages with limited training data, requires systematic evaluation.

Evaluating this technology is an ongoing need given its rapid evolution. Recent survey data show that 57% of U.S. physicians are already using or planning to adopt AI translation services within the next year, a faster adoption rate than any other AI use case surveyed.14 This creates urgency: clinicians are deploying tools with variable performance across languages, yet lack systematic data to guide their use. Physicians and patients require timely data on translation performance to inform patient education strategies. Such evaluations must assess performance across diverse languages, including those that have historically been underserved by both professional translation services and natural language processing research.

Traditional translation evaluation requires a bilingual human expert for each language pair, thereby limiting scalability. Recent advances in multilingual natural language processing (NLP) have produced automated metrics (including LaBSE, COMET, and BERTScore) that capture semantic similarity across languages.15-17 These metrics enable evaluation without requiring human evaluators for each language pair, making them particularly valuable for assessing low-resource languages where expert evaluators are scarce.

This study reports a systematic evaluation of LLM translation performance across eight languages using a multi-layer validation framework, focusing on English-to-target translation to reflect clinical scenarios where English-language materials must be made accessible to patients with non-English language preference. Because no single automated metric can fully characterize translation quality, we designed five successive validation layers, each addressing a specific limitation of the preceding approach: back-translation fidelity, comparison to professional translations, cross-model back-translation sensitivity analysis, inter-model translation concordance, and lexical borrowing quantification. We addressed six research questions: (1) Do LLM translations preserve medical meaning across high- and low-resource languages? (2) Does translation fidelity differ by language resource level? (3) How do LLM translations compare to professional translations? (4) Are back-translation fidelity results robust to the choice of back-translating model? (5) Do independently trained models converge on similar translations? (6) Do high fidelity scores for low-resource languages reflect genuine translation or retention of English terminology? We hypothesized that current-generation LLMs maintain translation fidelity across both high- and low-resource languages, achieving semantic preservation comparable to professional translation, with results that are robust to the choice of back-translating model, consistent across independently trained models, and not attributable to lexical borrowing of English medical terminology.

## Results

We completed 702 of 704 translation pairs (99.7%). Two Kimi K2 translations failed due to content filtering, likely triggered by anatomical or disease-related terminology in the source documents. While the failure rate was low (0.3%), content filtering by commercial LLM APIs represents a usability barrier for medical translation at scale, particularly for documents addressing sensitive health topics such as reproductive cancers or sexually transmitted infections.

### Methodological Validation

Back-translation of professional translations yielded high fidelity with original English (LaBSE: 0.92-0.94; BLEU: 43.9-55.8; Supplementary Table S2), establishing the benchmark against which to contextualize LLM performance.

### LLM Back-Translation Fidelity

All models achieved high semantic preservation (Table 1). Kruskal-Wallis testing revealed significant differences between models for both LaBSE ($H = 156.67$, $p < 0.001$) and BLEU ($H = 170.69$, $p < 0.001$). Dunn's post-hoc tests with Bonferroni correction showed Claude Opus 4.5 significantly outperformed all other models on semantic preservation (all pairwise $p < 0.001$). GPT-5.1 significantly outperformed Kimi K2 ($p = 0.003$) and Gemini 3 Pro ($p < 0.001$), while Gemini 3 Pro and Kimi K2 showed no significant difference ($p = 1.0$).

**Table 1. Back-Translation Fidelity by Model**

| Model | n | LaBSE (Mean +/- SD) | BLEU (Mean +/- SD) |
| --- | --- | --- | --- |
| Claude Opus 4.5 | 176 | 0.987 +/- 0.013 | 68.7 +/- 8.3 |
| GPT-5.1 | 176 | 0.957 +/- 0.045 | 64.3 +/- 7.4 |
| Kimi K2 | 174 | 0.940 +/- 0.053 | 54.9 +/- 8.7 |
| Gemini 3 Pro | 176 | 0.921 +/- 0.065 | 61.5 +/- 9.3 |

*Kruskal-Wallis: LaBSE $H = 156.67$, $p < 0.001$; BLEU $H = 170.69$, $p < 0.001$.*

### Language Performance by Resource Level

Low-resource languages achieved semantic preservation that did not differ significantly from that of high-resource languages (Figure 1, Supplementary Table S3). Tagalog (0.950) and Haitian Creole (0.955) scored on par with Spanish (0.954) and Vietnamese (0.953). Mann-Whitney U testing confirmed no significant difference between resource groups for semantic preservation (LaBSE: 0.948 vs 0.952, $p = 0.066$). This finding represents absence of a detected difference, not demonstrated equivalence; formal non-inferiority testing with a pre-specified clinically meaningful delta would be needed to establish equivalence, and the study may be underpowered to detect small differences in quality.

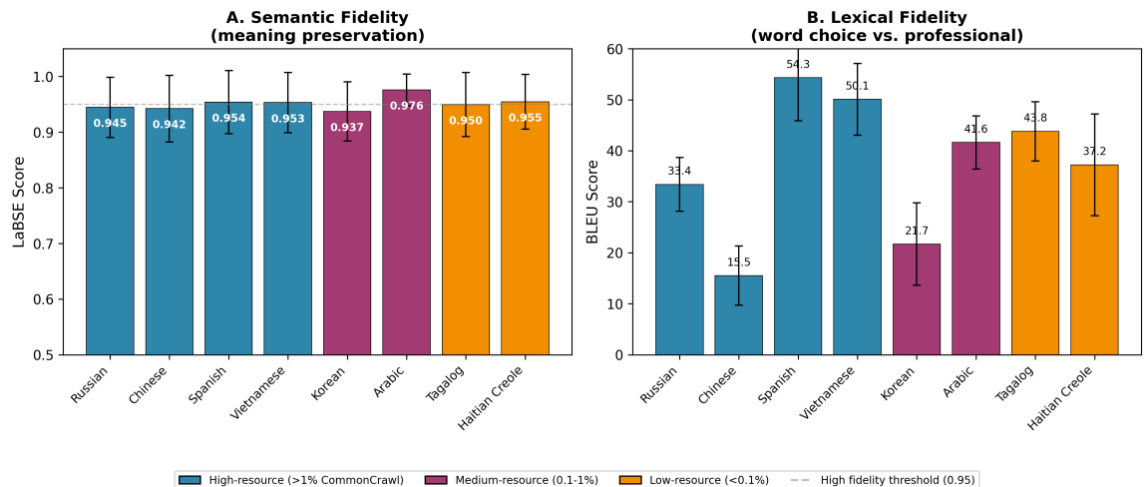

**Figure 1.** Semantic vs Lexical Fidelity by Language and Resource Level. (a) Semantic fidelity (LaBSE score) shows consistent preservation of meaning across languages, with low-resource languages (Tagalog, Haitian Creole) achieving scores that do not differ significantly from those of high-resource languages. The dashed line indicates a 0.95 high-fidelity threshold. (b) Lexical fidelity (BLEU score vs. professional translation) varies substantially across languages, reflecting morphological and syntactic differences rather than translation quality. Colors indicate resource level: blue = high-resource (greater than 1% CommonCrawl), purple = medium-resource (0.1-1%), orange = low-resource (less than 0.1%).

Although semantic preservation did not differ significantly across resource levels, high-resource languages showed better alignment with professional translations (COMET: 0.879 vs 0.837, $p < 0.001$). BLEU scores varied substantially by language (range: 15.5-54.3), while semantic metrics remained consistently high across all languages (LaBSE range: 0.937-0.976). The low BLEU for Chinese (15.5) reflects a known limitation of n-gram overlap metrics for logographic scripts: Chinese characters map to English words at a fundamentally different granularity, producing low lexical overlap even when meaning is fully preserved. This is precisely why semantic embedding metrics (LaBSE, COMET) are preferred for cross-lingual evaluation across typologically diverse languages.

### LLM vs Professional Translation

LLM translations approached professional quality across all models (Table 2). Kruskal-Wallis testing showed no significant differences between models for COMET (H = 2.07, p = 0.56) or BLEU (H = 5.50, p = 0.14). BERTScore showed a significant effect (H = 13.58, p = 0.004), with post-hoc tests revealing Claude Opus 4.5 significantly outperformed GPT-5.1 (p = 0.04) and Kimi K2 (p = 0.004).

**Table 2. LLM vs Professional Translation Quality**

| Model | n | BLEU | BERTScore | COMET |
| --- | --- | --- | --- | --- |
| Gemini 3 Pro | 176 | 39.4 +/- 14.9 | 0.845 +/- 0.060 | 0.876 +/- 0.054 |
| Claude Opus 4.5 | 176 | 37.3 +/- 14.4 | 0.859 +/- 0.043 | 0.874 +/- 0.054 |
| GPT-5.1 | 176 | 36.0 +/- 13.8 | 0.844 +/- 0.052 | 0.871 +/- 0.052 |
| Kimi K2 | 174 | 36.0 +/- 14.3 | 0.840 +/- 0.052 | 0.872 +/- 0.052 |

*Kruskal-Wallis: COMET H = 2.07, p = 0.56; BLEU H = 5.50, p = 0.14; BERTScore H = 13.58, p = 0.004.*

Cancer education materials achieved higher back-translation fidelity (mean LaBSE: 0.984) than Vaccine Information Statements (mean LaBSE: 0.919; Mann-Whitney U, p < 0.001).

### Cross-Model Sensitivity Analysis

Cross-model back-translation yielded results nearly identical to same-model back-translation (Table 3). Across 1,550 matched pairs, the mean delta between cross-model and same-model LaBSE was -0.0009 (+/- 0.020), indicating that same-model circularity does not meaningfully inflate fidelity scores. Claude Opus 4.5 forward translations showed a small decrease when back-translated by other models (delta = -0.011), while Kimi K2 and Gemini 3 Pro forward translations showed small increases under cross-model conditions (delta = +0.004 and +0.003, respectively), directly contradicting what circularity would predict.

The equity finding was preserved under cross-model conditions: low-resource languages achieved cross-model LaBSE of 0.948, which did not differ significantly from high-resource languages at 0.945 (Kruskal-Wallis H = 4.95, p = 0.084).

**Table 3. Cross-Model Back-Translation Sensitivity Analysis**

| Forward Model | Same-Model LaBSE | Cross-Model LaBSE | Delta |
|---|---|---|---|
| Claude Opus 4.5 | 0.987 | 0.976 | -0.011 |
| GPT-5.1 | 0.956 | 0.955 | -0.002 |
| Kimi K2 | 0.940 | 0.943 | +0.004 |
| Gemini 3 Pro | 0.921 | 0.924 | +0.003 |
| **Overall** | **0.950** | **0.949** | **-0.001** |

*1,550 matched same-model vs cross-model pairs. Positive delta indicates higher cross-model scores, contradicting the circularity hypothesis.*

### Inter-Model Translation Concordance

Across 1,047 pairwise comparisons between translations produced by independently trained models, overall concordance was high (LaBSE: 0.946 +/- 0.056; BERTScore F1: 0.892 +/- 0.047; Table 4). Arabic translations showed the highest concordance (LaBSE: 0.973), while Haitian Creole showed the lowest (LaBSE: 0.934). By resource level, medium-resource languages showed the highest concordance (LaBSE: 0.956), followed by high-resource (0.945) and low-resource (0.937); this difference was statistically significant (Kruskal-Wallis H = 16.75, p < 0.001), though all three resource levels exceeded the 0.90 threshold for high semantic preservation.

Among model pairs, Gemini 3 Pro and Kimi K2 showed the highest concordance (LaBSE: 0.963), while Claude Opus 4.5 and Gemini 3 Pro showed the lowest (LaBSE: 0.925). Cancer education materials showed higher inter-model concordance (LaBSE: 0.982) than Vaccine Information Statements (LaBSE: 0.909), consistent with the back-translation findings.

**Table 4. Inter-Model Translation Concordance by Language**

| Language | Resource Level | N Pairs | LaBSE (Mean +/- SD) | BERTScore F1 (Mean +/- SD) |
|---|---|---|---|---|
| Arabic | Medium | 126 | 0.973 +/- 0.019 | 0.917 +/- 0.022 |
| Spanish | High | 132 | 0.951 +/- 0.060 | 0.917 +/- 0.055 |
| Vietnamese | High | 132 | 0.948 +/- 0.054 | 0.897 +/- 0.047 |
| Russian | High | 132 | 0.943 +/- 0.061 | 0.897 +/- 0.044 |
| Tagalog | Low | 129 | 0.941 +/- 0.057 | 0.878 +/- 0.040 |
| Korean | Medium | 132 | 0.940 +/- 0.048 | 0.878 +/- 0.036 |
| Chinese | High | 132 | 0.939 +/- 0.062 | 0.872 +/- 0.055 |
| Haitian Creole | Low | 132 | 0.934 +/- 0.061 | 0.877 +/- 0.040 |

| Resource Level | N Pairs | LaBSE (Mean +/- SD) | BERTScore F1 (Mean +/- SD) |
|---|---|---|---|
| Medium | 258 | 0.956 +/- 0.040 | 0.897 +/- 0.036 |
| High | 528 | 0.945 +/- 0.060 | 0.896 +/- 0.053 |
| Low | 261 | 0.937 +/- 0.059 | 0.878 +/- 0.040 |
| **Overall** | **1,047** | **0.946 +/- 0.056** | **0.892 +/- 0.047** |

*Concordance is computed as pairwise LaBSE similarity and BERTScore F1 between all model translations for each document-language pair (6 comparisons per group). Languages sorted by LaBSE descending.*

### Lexical Borrowing Analysis

Lexical borrowing rates varied substantially across languages (Table 5). Tagalog retained the highest proportion of English medical terms (68.2% +/- 17.2%), followed by Vietnamese (25.1% +/- 12.6%), Haitian Creole (24.3% +/- 14.0%), Korean (22.2% +/- 13.8%), Spanish (21.2% +/- 11.4%), Arabic (20.2% +/- 12.8%), Chinese (19.1% +/- 10.8%), and Russian (16.7% +/- 9.4%). Borrowing rates differed significantly across resource levels (Kruskal-Wallis H = 120.03, p < 0.001), with low-resource languages showing the highest rates (mean: 46.3%) compared to medium-resource (21.2%) and high-resource (20.5%) languages.

Critically, borrowing rate showed no significant correlation with LaBSE scores in low-resource languages (Spearman rho = +0.018, p = 0.82, n = 175). This finding directly addresses the concern that high LaBSE scores for Tagalog and Haitian Creole reflect verbatim retention of English terminology rather than genuine semantic preservation. When examined by individual language, the correlation was negative and significant for Haitian Creole (rho = -0.380, p = 0.0003), meaning that translations retaining more English terms actually scored lower on semantic fidelity. The overall correlation across all languages was also negative (rho = -0.219, p < 0.001), indicating that lexical borrowing tends to reduce rather than inflate fidelity scores.

Among models, GPT-5.1 retained a higher proportion of English medical terms (34.6%) compared to Gemini 3 Pro (26.8%), Claude Opus 4.5 (24.0%), and Kimi K2 (23.1%), though this variation did not correspond to differences in overall translation quality. For Haitian Creole, French medical term borrowing was minimal (mean: 1.0%), suggesting that the models predominantly translate into the English-derived register of Haitian Creole medical vocabulary rather than the French-derived register.

**Table 5. Lexical Borrowing of English Medical Terminology**

| Language | Resource Level | Borrowing Rate (Mean +/- SD) | Spearman rho vs LaBSE | p-value |
|---|---|---|---|---|
| Tagalog | Low | 0.682 +/- 0.172 | +0.189 | 0.079 |
| Vietnamese | High | 0.251 +/- 0.126 | -0.329 | 0.002 |
| Haitian Creole | Low | 0.243 +/- 0.140 | -0.380 | 0.0003 |
| Korean | Medium | 0.222 +/- 0.138 | -0.350 | 0.001 |
| Spanish | High | 0.212 +/- 0.114 | -0.583 | < 0.001 |
| Arabic | Medium | 0.202 +/- 0.128 | -0.089 | 0.414 |
| Chinese | High | 0.191 +/- 0.108 | -0.432 | < 0.001 |
| Russian | High | 0.167 +/- 0.094 | -0.375 | 0.0003 |

| Group | n | Borrowing Rate | Spearman rho vs LaBSE | p-value |
|---|---|---|---|---|
| Low-resource combined | 175 | 0.463 | +0.018 | 0.817 |
| Overall | 702 | 0.271 | -0.219 | < 0.001 |

*Borrowing rate is the proportion of English medical terms (from a vocabulary of 213 terms) retained verbatim in each translation. Languages sorted by borrowing rate descending. Kruskal-Wallis across resource levels: H = 120.03, p < 0.001.*

## Discussion

This study applied a multi-layer validation framework to evaluate LLM medical translation across eight languages, with each successive layer designed to address a limitation of the preceding approach. The five layers produced converging results with important implications for health equity. Back-translation fidelity showed that all frontier LLMs reliably preserved medical meaning (LaBSE greater than 0.92), with no significant difference by language resource level (p = 0.066). Comparison to professional translations confirmed that LLM output approached professional quality (COMET: 0.87-0.88) without significant inter-model differences. Cross-model sensitivity analysis ruled out same-model circularity as a driver of these results (delta = -0.0009). Inter-model concordance demonstrated that four independently trained models converge on similar translations (LaBSE: 0.946), providing convergent validity independent of back-translation entirely. Finally, lexical borrowing analysis confirmed that high fidelity scores for low-resource languages reflect genuine semantic preservation rather than English term retention.

### Key Findings

**Equivalent fidelity for low-resource languages.** Tagalog and Haitian Creole (languages comprising less than 0.01% of CommonCrawl) achieved semantic similarity scores that were not statistically distinguishable from those of Spanish and Vietnamese (p = 0.066). This represents a meaningful advance: earlier evaluations of LLMs demonstrated significant performance degradation for digitally underrepresented languages.18 A 2024 study found that ChatGPT and Google Translate had significantly more clinically significant errors for Haitian Creole compared to Spanish and Portuguese when translating pediatric discharge instructions.13 Our finding of equivalent performance for Haitian Creole suggests frontier models released in 2025 may have substantially improved multilingual capabilities for formal patient education materials.

However, lexical overlap with professional translations (BLEU) was lower for low-resource languages, indicating that while LLMs accurately convey the information, they may phrase it differently from a human translator. The likely explanation is that professional translation conventions are more established for high-resource languages such as Spanish and Chinese, providing LLMs with more examples to learn from during training.

**Scalable evaluation methodology.** Automated multilingual metrics enabled rigorous comparison across eight languages without requiring bilingual experts for each pair, a practical necessity when expert evaluators are scarce or unavailable for low-resource languages or for the volume of clinical documentation required to fully engage patients in their care.

**Semantic preservation is more consistent than lexical overlap.** While BLEU scores varied widely by language (range: 15.5-54.3, reflecting morphological and syntactic differences), semantic metrics remained consistently high across all languages (LaBSE range: 0.937-0.976). This validates the importance of using multilingual embedding-based metrics rather than relying solely on n-gram overlap for evaluating translation quality across typologically diverse languages.

**Model convergence.** All four frontier models performed within a narrow band, suggesting that model selection is less critical than language-pair selection. This convergence may reflect shared training approaches, similar underlying architectures, or saturation of translation capabilities at the frontier. Claude Opus 4.5 achieved the highest semantic preservation through back-translation, while Gemini 3 Pro led on lexical agreement with professional translations. These complementary strengths suggest potential for ensemble approaches in production systems.

### Addressing the Circularity Concern

A legitimate methodological concern with back-translation evaluation is that using the same LLM for both forward and back translation could inflate fidelity scores if the model exhibits consistent translation biases in both directions. Our cross-model sensitivity analysis directly addresses this concern. Across 1,550 matched pairs, cross-model back-translation scores were nearly identical to same-model scores (delta = -0.0009). Two of four forward-translating models actually showed higher scores under cross-model conditions, the opposite of what circularity would predict. The equity finding was also preserved: low-resource languages achieved cross-model LaBSE of 0.948, not significantly different from high-resource languages (p = 0.084).

### Convergent Validity from Inter-Model Concordance

The inter-model concordance analysis provides a validation pathway that is entirely independent of back-translation. By comparing translations produced by four independently trained models (developed by four different organizations using different training data, architectures, and optimization procedures), we can assess whether the models converge on similar translations without any back-translation step. The high overall concordance (LaBSE: 0.946) indicates that these models produce translations that are semantically very similar to one another. If any single model were producing systematically biased or incorrect translations, its outputs would diverge from those of the other models, resulting in lower concordance scores.

The convergence pattern across model pairs is informative. The highest concordance was observed between Gemini 3 Pro and Kimi K2 (LaBSE: 0.963), while the lowest was between

Claude Opus 4.5 and Gemini 3 Pro (LaBSE: 0.925). Even this lowest concordance exceeds the 0.90 threshold for high semantic preservation, indicating substantial agreement across all model pairs. The concordance between independently trained multilingual models suggests that these models have learned a shared representation of medical translation that converges on accurate output, consistent with research on cross-lingual representation learning.19

Cancer education materials showed notably higher concordance (LaBSE: 0.982) than Vaccine Information Statements (LaBSE: 0.909), paralleling the back-translation results and suggesting that the simpler, more declarative language of cancer education materials permits greater model agreement.

### Interpreting the Low-Resource Language Finding

The strong performance of Tagalog and Haitian Creole on semantic fidelity metrics warrants careful interpretation. A plausible alternative explanation is that high LaBSE scores for these languages reflect lexical borrowing from English rather than genuine translation. Medical Tagalog frequently retains English terminology (e.g., "chemotherapy," "vaccine," "cancer"), as do many post-colonial languages in technical domains.20 If models merely retain English medical terms rather than translating them, back-translation fidelity could be artificially inflated.

Our lexical borrowing analysis directly tests this hypothesis with quantitative data. Tagalog does indeed retain a high proportion of English medical terms (68.2%), substantially more than any other language in our study. However, the critical finding is that borrowing rate shows no significant correlation with LaBSE scores in low-resource languages combined (rho = +0.018, p = 0.82). When examined individually, Haitian Creole shows a significant negative correlation between borrowing rate and LaBSE (rho = -0.380, p = 0.0003), meaning that translations with more retained English terms actually achieve lower semantic fidelity scores. The overall correlation across all languages is also negative (rho = -0.219, p < 0.001).

These findings suggest that lexical borrowing does not inflate LaBSE scores. Indeed, the negative correlations observed in most languages indicate that retaining English terms may slightly reduce fidelity, perhaps because verbatim retention disrupts the semantic coherence of the target-language sentence as encoded by LaBSE. The high semantic fidelity scores for Tagalog and Haitian Creole appear to reflect genuine translation quality rather than an artifact of English term retention.

It is worth noting that LLMs exhibit bias toward loanwords and may struggle to distinguish borrowed from native vocabulary,21 which could affect translation strategies for languages with extensive borrowing patterns. GPT-5.1 retained substantially more English terms (34.6%) than other models (23-27%), suggesting that models differ in their propensity for lexical borrowing, though this variation did not correspond to differences in translation quality.

### Clinical Implications

These findings have implications for the delivery of equitable healthcare at scale. Medical translation services are often unavailable or delayed for speakers of less common languages, creating barriers to informed healthcare decision-making. Our multi-method validation results suggest that for formal patient education materials, LLMs can reliably convey medical information to Tagalog and Haitian Creole speakers with meaning preservation comparable to that achieved for Spanish.

Recent HHS guidance on Section 1557 acknowledges that "exigent circumstances" may arise where machine translation is used before qualified human review is feasible, provided that "the machine translation must be subsequently checked by a qualified human translator as soon as practicable."22 Our findings provide empirical support for such use: frontier LLMs maintain semantic fidelity across resource levels for the types of standardized materials studied here.

### Policy Implications

Current federal rules implementing Section 1557 require that machine translations of critical medical documents be reviewed by a "qualified human translator."6 While intended to ensure quality, this requirement may inadvertently limit access for speakers of low-resource languages where qualified translators are scarce or unavailable. Our findings suggest that for standardized patient education materials, frontier LLMs achieve semantic fidelity that does not differ significantly across resource levels. Whether performance-based criteria might better balance quality assurance with equitable access is a question that warrants further investigation, including patient comprehension studies and evaluation of additional document types, before any policy changes could be considered.

### Limitations

**Potential training data contamination.** The medical documents evaluated (CDC Vaccine Information Statements, American Cancer Society patient education materials) are publicly available resources that may be present in LLM training corpora. This potential contamination could artificially inflate performance if models memorized specific content rather than developing generalizable translation capabilities. The comparable performance of low-resource languages raised concern about this possibility. To address this, we conducted a sentence-reordering sensitivity analysis (Supplementary Analysis S1): we randomly shuffled sentence order within documents and repeated the translation pipeline. The hypothesis was that memorized documents would show significant performance degradation when the structure was disrupted. Results showed no significant change in 83% of model-language combinations, suggesting our results primarily reflect translation capability rather than memorization. However, Claude Opus 4.5 showed a 6.0% decrease for Tagalog when sentence order was shuffled ($p = 0.041$), which may indicate partial memorization for this specific model-language pair and warrants caution in interpreting its Tagalog scores.

**Automated metrics only.** While we employed multiple validated metrics, automated evaluation cannot fully capture the nuances of medical terminology, cultural appropriateness, or potential for patient misunderstanding. Semantic embedding metrics may miss clinically critical errors such as hallucinated dosages, reversed polarities (e.g., "take" vs. "do not take"), or subtle shifts in tone that affect patient compliance. Future work should incorporate bilingual clinician review for a randomized subset of translations to validate automated metrics against clinical safety assessments.

**Limited document types.** Our corpus consisted solely of patient education materials. Results may not generalize to clinical notes, consent forms, or other medical document types with different linguistic characteristics.

**Single translation direction.** We evaluated English-to-target translation only. Target-to-English translation (relevant for patient-provider communication) may show different patterns.

**Arabic dialect limitation.** Arabic was evaluated without distinguishing Modern Standard Arabic (MSA) from dialectal varieties. MSA performance may not generalize to Moroccan Darija, Levantine, Egyptian, or Gulf dialects, which represent the majority of Arabic-speaking patients in U.S. healthcare settings.

**Snapshot evaluation.** LLM capabilities evolve rapidly; these results reflect model versions available in late 2025 and may not reflect future or past capabilities.

## Future Directions

Several extensions of this work warrant investigation. First, evaluation should be expanded to include additional clinical document types, such as clinician notes, medication instructions, and informed consent forms. Second, given the rapid evolution of LLMs, longitudinal assessment is needed to track translation performance as models are updated. Third, ensemble approaches should be tested to determine whether combining outputs from multiple models yields more reliable translations than a single model. Fourth, evaluation should extend to real-world understanding by patients of medical information and whether these efforts enhance their ability to understand and participate fully in their care. Fifth, the technology should be further assessed on the ability to not just translate paper documents, but also to enhance meaningful patient education by providing information in a specified language, at a desired reading level, and in oral or video formats. Sixth, given the limitations identified above, the data from this study cannot support relaxing human review requirements for machine translation of critical medical documents; patient comprehension studies and assessment of clinical outcomes would be necessary before any such policy changes could be considered.

## Conclusion

Frontier LLMs reliably preserve medical meaning through translation across both high-resource and low-resource languages. Critically, low-resource languages achieve semantic preservation that does not differ significantly from that of high-resource languages, suggesting that LLMs could extend medical translation access to historically underserved populations. Multiple validation approaches, including cross-model back-translation, inter-model concordance, and lexical borrowing analysis, converge on this finding, providing robust evidence that the observed translation quality is genuine rather than an artifact of evaluation methodology.

These findings suggest significant advancement in LLM performance in medical translation of formal patient education information. Should continued assessment show similar performance for other types of medical information, machine translation may open new pathways for patient education in clinical settings.

# Methods

## Study Design

We conducted a cross-sectional evaluation of four frontier LLMs for medical document translation across eight languages, using automated metrics and validation against professional translation baselines.

**Models:** GPT-5.1 (OpenAI, 2025), Claude Opus 4.5 (Anthropic, 2025), Gemini 3 Pro (Google, 2025), and Kimi K2 Thinking (Moonshot AI, 2025). All models were accessed via their respective APIs using default parameters.

**Languages:** Russian, Chinese (Simplified), Spanish, Vietnamese, Korean, Arabic, Tagalog, and Haitian Creole, representing diverse linguistic families, scripts, and resource levels.

## Data Sample

We assembled 22 medical education documents from two domains:

**Vaccine Information Statements (n=11):** CDC documents distributed via Immunize.org covering Hepatitis B, HPV, Influenza, MMR, Meningococcal ACWY, Pneumococcal vaccines, Polio, Shingles, Tdap, and Varicella.23

**Cancer Education Materials (n=11):** American Cancer Society documents covering post-diagnosis guidance (breast, cervical, colorectal, lung, prostate cancer), treatment side effects, and skin cancer resources.24

Documents were selected based on the availability of professional translations in all eight target languages.

## Language Classification

Languages were classified by CommonCrawl representation, a primary LLM training corpus:18,25 high-resource (greater than 1%: Spanish, Chinese, Russian, Vietnamese), medium-resource (0.1-1%: Korean, Arabic), and low-resource (less than 0.1%: Tagalog, Haitian Creole). See Supplementary Table S1 for detailed classification criteria including CommonCrawl percentages and U.S. speaker populations.

## Translation Pipeline

For each document-language-model combination:

1. Forward Translation: English to target language (LLM)
2. Back-Translation: Target language to English (same LLM)
3. Professional Back-Translation: Professional translation to English (same LLM)

Forward and back translation used separate API calls, ensuring that back translation had no access to the original English text. This yielded 704 translation pairs (22 documents x 8 languages x 4 models).

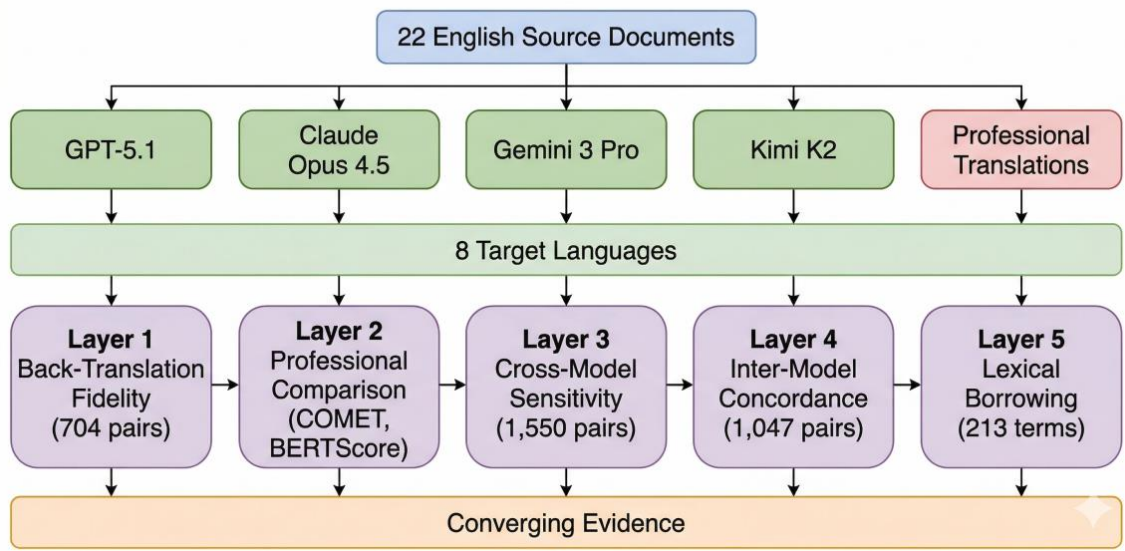

**Figure 2.** Multi-Layer Validation Framework for LLM Medical Translation. Twenty-two English source documents were translated into 8 target languages by four frontier LLMs (GPT-5.1, Claude Opus 4.5, Gemini 3 Pro, Kimi K2). Translation quality was assessed through five successive validation layers: Layer 1, back-translation fidelity (704 pairs); Layer 2, comparison to professional human translations (COMET, BERTScore); Layer 3, cross-model back-translation sensitivity analysis (1,550 pairs); Layer 4, inter-model translation concordance (1,047 pairwise comparisons); Layer 5, lexical borrowing quantification (213 terms).

**Multi-Method Validation Framework**

No single automated metric can fully characterize translation quality, and each evaluation approach carries inherent limitations. Back-translation fidelity, while aligned with established cross-cultural adaptation frameworks,[26,27] may be influenced by shared model biases when the same system performs both forward and back translation. Comparison to professional translation provides an external reference standard but is constrained to the subset of languages and documents for which professional translations exist. We therefore designed a multi-layer validation framework in which each successive analysis addresses a specific limitation of the preceding approach, building a converging evidence base rather than relying on any single metric.

Translation quality can be assessed through two complementary lenses: *lexical fidelity*, which measures whether the same words and phrases are used, and *semantic fidelity*, which measures whether the meaning is preserved regardless of exact wording. High lexical fidelity implies high semantic fidelity, but the reverse is not true: a translation can preserve meaning while using different vocabulary. For example, "Take medication twice daily" and "Take medicine two times each day" have imperfect lexical overlap but identical meaning. This distinction is particularly important for cross-lingual evaluation, as languages with different scripts or morphological structures (e.g., Chinese, Russian) will inherently show lower lexical overlap even when meaning is fully preserved.

The five validation layers are as follows:

**Layer 1: Back-Translation Fidelity.** As a first-order assessment, we compared back-translations to original English text using BLEU (lexical overlap, 0-100 scale; scores above 50 indicate strong word-level agreement),28 LaBSE (semantic similarity, 0-1 scale),15 XLM-RoBERTa,29 and mBERT.30 LaBSE scores greater than 0.90 indicate high semantic preservation; differences of 0.05 represent meaningful variation. To validate this methodology, we also back-translated professional translations through each LLM; high fidelity scores on this benchmark confirm that back-translation meaningfully reflects translation quality.

**Layer 2: Comparison to Professional Translation.** Because back-translation evaluates round-trip consistency rather than accuracy against a known-correct reference, we next compared LLM translations directly to professional translations using BLEU, chrF,31 BERTScore,16 and COMET.17 This provides an external reference standard, independent of the back-translation pathway.

**Layer 3: Cross-Model Sensitivity Analysis.** A remaining concern is that using the same LLM for both forward and back translation could inflate fidelity scores through shared translation biases (circularity). To test this, we performed back-translation of each forward translation using all other available models, yielding 1,550 matched same-model versus cross-model comparison pairs across 9 model pairings. Kimi K2 was included as a forward translator but excluded as a back-translator due to processing constraints and content filtering limitations. If circularity were meaningfully inflating scores, cross-model back-translation would yield systematically lower fidelity than same-model back-translation.

**Layer 4: Inter-Model Translation Concordance.** To establish convergent validity entirely independent of back-translation, we compared translations produced by different models head-to-head. For each of the 176 document-language groups (22 documents x 8 languages), we computed pairwise LaBSE similarity and BERTScore F1 between all model translation pairs (4 choose 2 = 6 comparisons per group), yielding 1,047 total pairwise comparisons. Four independently trained models, developed by four different organizations (OpenAI, Anthropic, Google, Moonshot AI) using different training data, architectures, and optimization procedures, serve as independent replicates: high concordance indicates that translation quality reflects the underlying content rather than any single model's idiosyncrasies.

**Layer 5: Lexical Borrowing Quantification.** Finally, we tested whether high semantic fidelity scores for low-resource languages reflect genuine translation or retention of English medical terminology (lexical borrowing). Many post-colonial languages retain English or French loanwords for technical vocabulary, which could inflate similarity metrics without reflecting translation capability. We compiled a vocabulary of 213 English medical terms from the source documents (Supplementary Data 1) and computed, for each of the 704 translations, the proportion of these terms retained verbatim (case-insensitive) in the target-language output. For Haitian Creole, which derives substantially from French, we additionally compiled 175 French medical terms and computed French borrowing rates. Spearman rank correlations between borrowing rate and LaBSE scores were computed for each language individually and for low-resource languages combined. If lexical borrowing inflates LaBSE scores, a positive correlation would be expected.

### Statistical Analysis

We report means and standard deviations. Model comparisons used Kruskal-Wallis tests with Dunn's post-hoc correction.32 Language comparisons stratified by resource level. Spearman rank correlations were used for borrowing rate versus LaBSE analyses. Significance was set at $p < 0.05$. Analyses used Python 3.11 with SciPy (v1.13.1) and statsmodels (v0.14.1).

### Data Availability

All source documents, translation outputs, and evaluation metrics are available at: https://github.com/hallixrap/llm-translation-study

### Code Availability

All analysis code, including translation pipeline scripts, metric computation, cross-model sensitivity analysis, inter-model concordance analysis, and lexical borrowing quantification scripts, is available at: https://github.com/hallixrap/llm-translation-study

## Acknowledgements

We thank the CDC, Immunize.org, and the American Cancer Society for making professionally-translated health education materials publicly available.

Support for this research was provided by the Commonwealth Fund. The views presented here are those of the authors and should not be attributed to the Commonwealth Fund or its directors, officers, or staff.

## Author Contributions

C.A. conceived the study, developed the translation pipeline, performed all analyses, and drafted the manuscript. E.J.P.G. and J.L. contributed to data acquisition and analysis. T.K. provided statistical guidance. A.L. and N.S. contributed to data interpretation and manuscript revision. S.M. contributed to study design and data acquisition. J.C. and K.S. supervised the study and provided critical revision of the manuscript. All authors read and approved the final manuscript.

## Competing Interests

The authors declare no competing interests.

## Supplementary Materials

### Supplementary Table S1: Language Classification Details

| Language | Resource Level | CommonCrawl % | U.S. Speakers (millions) |
|---|---|---|---|
| Russian | High | 6.48 | 0.9 |
| Chinese (Simplified) | High | 6.18 | 3.5 |
| Spanish | High | 4.41 | 41.8 |
| Vietnamese | High | 1.08 | 1.6 |
| Korean | Medium | 0.80 | 1.1 |
| Arabic | Medium | 0.67 | 1.3 |
| Tagalog | Low | 0.008 | 1.8 |
| Haitian Creole | Low | 0.003 | 0.9 |

High-resource: greater than 1%; medium-resource: 0.1-1%; low-resource: less than 0.1%. U.S. speaker populations from American Community Survey.2

### Supplementary Analysis S1: Sentence Reordering Sensitivity Analysis

To address training data contamination concerns, we tested whether performance reflects document memorization versus true translation capability by randomly shuffling sentence order in 10 documents and comparing performance.

**Results:** No significant performance change in 10 of 12 model-language combinations (83%). Two comparisons reached significance: Claude Opus 4.5 with Tagalog showed 6.0% decrease (p = 0.041); Gemini 3 Pro with Tagalog showed 3.0% increase (p = 0.026), opposite to what memorization would predict.

**Negative Control:** A direct comparison of the original and shuffled documents (without translation) yielded a LaBSE of 0.81, confirming that the metric is sensitive to structural changes.

Back-translations from both conditions achieved approximately 0.95, well above this baseline, indicating semantic preservation rather than memorization.

### Supplementary Table S2: Professional Translation Back-Translation Fidelity

| Model | BLEU | BERTScore | LaBSE |
|---|---|---|---|
| GPT-5.1 | 55.8 | 0.919 | 0.940 |
| Claude Opus 4.5 | 54.7 | 0.924 | 0.942 |
| Gemini 3 Pro | 48.3 | 0.913 | 0.928 |
| Kimi K2 | 43.9 | 0.912 | 0.934 |

### Supplementary Table S3: Translation Fidelity by Language

| Language | Resource | Back-Translation LaBSE | vs Professional BLEU |
|---|---|---|---|
| Arabic | Medium | 0.976 | 41.6 |
| Haitian Creole | Low | 0.955 | 37.2 |
| Spanish | High | 0.954 | 54.3 |
| Vietnamese | High | 0.953 | 50.1 |
| Tagalog | Low | 0.950 | 43.8 |
| Russian | High | 0.945 | 33.4 |
| Chinese | High | 0.942 | 15.5 |
| Korean | Medium | 0.937 | 21.7 |

### Supplementary Data 1: English Medical Term Vocabulary

The complete list of 213 English medical terms used for lexical borrowing quantification is provided as Supplementary Data 1 (CSV file). Terms were compiled from the 22 source documents and categorized into three domains: Vaccines and Immunization (63 terms), Cancer and Oncology (78 terms), and General Medical (72 terms). Terms were matched case-insensitively in target-language translations.

### Supplementary Data 2: French Medical Term Vocabulary

The complete list of 175 French medical terms used for Haitian Creole French-borrowing analysis is provided as Supplementary Data 2 (CSV file).